%% file: arXiv.tex
\documentclass[10pt,twocolumn,letterpaper]{article}

\usepackage{cvpr}
\usepackage{times}
\usepackage{epsfig}
\usepackage{graphicx}
\usepackage{amsmath}
\usepackage{amssymb}
\usepackage{subfig}
\usepackage{authblk}
\usepackage[symbol]{footmisc}

\newcommand\blfootnote[1]{%
	\begingroup
	\renewcommand\thefootnote{}\footnote{#1}%
	\addtocounter{footnote}{-1}%
	\endgroup
}
\newcommand\FirstEmailMark{\footnotemark[2]}
\newcommand\SecondEmailMark{\footnotemark[3]}
\newcommand\thirdEmailMark{\footnotemark[1]}

\usepackage[pagebackref=true,breaklinks=true,letterpaper=true,colorlinks,bookmarks=false]{hyperref}
\cvprfinalcopy 

\def\cvprPaperID{6264} 

\ifcvprfinal\fi
\begin{document}
\author[1, 2]{Xudong Liu \protect\FirstEmailMark }

\author[1]{Ruizhe Wang \protect\SecondEmailMark }
\author[1]{Chih-Fan Chen \protect\SecondEmailMark}
\author[2]{Minglei Yin \protect\FirstEmailMark}
\author[1]{Hao Peng \protect\SecondEmailMark}
\author[1]{Shukhan Ng \protect\SecondEmailMark}
\author[2]{Xin Li \protect\thirdEmailMark}

\affil[1]{Oben, Inc}
\affil[2]{West Virginia University}

    \title{Face Beautification: Beyond Makeup Transfer}
    \twocolumn[{%
    \renewcommand\twocolumn[1][]{#1}%
    \maketitle
    \begin{center}
        \centering
        \includegraphics[width=1 \textwidth]{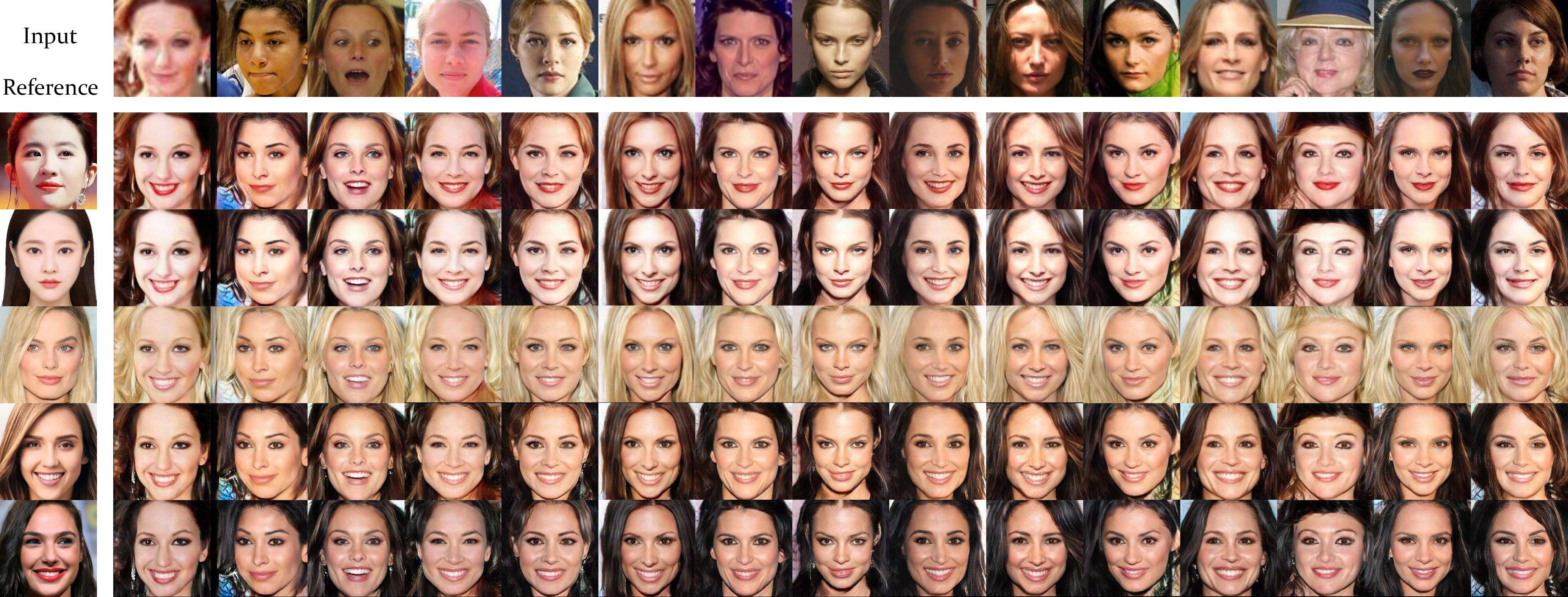}
        \captionof{figure}{Face beautification as many-to-many image translation: our approach integrates style-based beauty representation with beauty score prediction model and is capable of fine-granularity control.}
        \label{fig:teaser}
    \end{center}%
    }]
\maketitle


\begin{abstract}
\blfootnote{ $\dagger$ \{xdliu,my0033\}@mix.wvu.edu}
\blfootnote{ $\ddagger$ \{ruizhe, chihfan, hpeng,shukhan\}@oben.com}
\blfootnote{ $\ast$ xin.li@mail.wvu.edu}
Facial appearance plays an important role in our social lives. Subjective perception of women's beauty depends on various face-related (e.g., skin, shape, hair) and environmental (e.g., makeup, lighting, angle) factors. Similar to cosmetic surgery in the physical world, virtual face beautification is an emerging field with many open issues to be addressed. Inspired by the latest advances in style-based synthesis and face beauty prediction, we propose a novel framework of face beautification.  For a given reference face with a high beauty score, our GAN-based architecture is capable of translating an inquiry face into {\em a sequence of} beautified face images with referenced beauty style and targeted beauty score values. To achieve this objective, we propose to integrate both style-based beauty representation (extracted from the reference face) and beauty score prediction (trained on SCUT-FBP database) into the process of beautification. Unlike makeup transfer, our approach targets at many-to-many (instead of one-to-one) translation where multiple outputs can be defined by either different references or varying beauty scores. Extensive experimental results are reported to demonstrate the effectiveness and flexibility of the proposed face beautification framework.
\end{abstract}

\section{Introduction}
\input{Section/Introduction.tex}

\section{Related Works}
\input{Section/RelatedWork.tex}

\section{Proposed Method}
\input{Section/ProposedMethod.tex}

\section{Experimental Setup}
\input{Section/Experiments.tex}

\section{Experimental Results and Evaluations}

\input{Section/ResultsEvaluation.tex}

\section{Conclusions and Future Works}
\input{Section/Conclusion.tex}

\section{Appendix}
\input{Section/Supplemental.tex}

{\small
\bibliographystyle{ieee_fullname}
\bibliography{arXiv}
}

\end{document}

%% file: Section/Introduction.tex
Facial appearance plays an important role in our social lives \cite{bull2012social}. People with attractive faces have many advantages in their social activities such as dating and voting \cite{little2011facial}. It has been found that attractive people enjoy higher chances of getting dating \cite{riggio1984role}, and their partners are more likely to gain satisfaction when compared to dating with less attractive ones \cite{berscheid1971physical}. It has also been found that faces could affect hiring decisions and influence voting behavior \cite{little2011facial}. Overwhelmed by social fascination with beauty, women with unattractive faces may suffer from social isolation, depression, and even psychological disorders \cite{bull2012social,rankin1998quality,phillips1993body,macgregor1989social,bradbury1994psychology}. Consequently, there is strong demand for face beautification both in the physical world (e.g., facial makeup and cosmetic surgeries) and in the virtual space (e.g., beautification cameras and filters). 

The problem of face beautification has been extensively studied by philosophers, psychologists and plastic surgeons. Rapid advances in imaging technology and social media greatly expedited the popularity of digital photos especially selfies in our daily lives. Most recently, virtual face beautification based on the idea of makeup application or transfer has been developed in computer vision communities: PairedCycleGAN \cite{chang2018pairedcyclegan}, BeautyGAN\cite{li2018beautygan}, BeautyGlow \cite{chen2019beautyglow}. Although these existing works have achieved impressive results, we argue that face beautification based on makeup transfer only has fundamental limitations. Without changing important facial attributes (e.g., shape and lentigo), the application of makeup - abstracted by image-to-image translation \cite{zhu2017unpaired,huang2018multimodal,DRIT} - can only improve the beauty score to some extent. 

A more flexible and promising framework is to formalize the process of face beautification by {\em one-to-many} translation where the destination can be defined in many different manners. On one hand, we can target at producing a sequence of output images with monotonically increased beauty scores by gradually transferring the style-based beauty representation learned from a given reference (with a high beauty score). On the other hand, we can also produce a variety of personalized beautification results by learning from a sequence of references (e.g., celebrities with different beauty style). Under this framework, face beautification can be made more flexible - e.g., we can transfer the beauty style from a reference image to reach a specified beauty score, which is beyond the reach of makeup transfer \cite{li2018beautygan,chen2019beautyglow}. 

To achieve this objective, we propose a novel generative adversarial network (GAN)-based architecture in this paper.  Inspired by the latest advances in style-based synthesis (e.g., styleGAN\cite{karras2019style})  and  face beauty  understanding from data \cite{liu2019understanding}, we propose to integrate both style-based beauty representation (extracted  from  the  reference  face)  and beauty  score  prediction (trained on SCUT-FBP database \cite{xie2015scut}) into the process of face beautification. More specifically, style-based beauty representations will be learned from both inquiry and reference images first via light convolutional neural network (LightCNN) and leveraged to guide the process of style transfer (actual beautification). Then a dedicated GAN-based architecture integrated with reconstruction, beauty and identity loss functions is constructed. In order to have a fine-granularity control of the beautification process, we have invented a simple yet effective reweighting strategy of gradually improving the beauty score in synthesized images until reaching the target (specified by the reference image).


Our key contributions are summarized as follows:

\begin{itemize}
    \item A forward-looking view toward virtual face beautification and a holistic style-based approach beyond makeup transfer (e.g., BeautyGAN and BeautyGlow). We argue that facial beauty scores offer a quantitative solution to guiding the process of face beautification.
    \item A face beauty prediction network based on fine-tuning of LightCNN is trained and integrated into the proposed style-based face beautification network. The prediction module provides valuable feedback to the synthesis module while approaching the desirable beauty score.
    \item A piggyback trick to extract both identity and beauty features from fine-tuned LightCNN and design of loss functions reflecting the tradeoff between identity preservation and face beautification.
    \item To the best of our knowledge, this is the first work capable of delivering face beautification results with fine-granularity control (i.e., a sequence of face images approaching the reference one with monotonically increasing beauty scores). 
    \item A comprehensive evaluation shows the superiority of the proposed approach when compared to existing state-of-the-art image-to-image transfer techniques including CycleGAN \cite{zhu2017unpaired}, MUNIT\cite{huang2018multimodal}, and DRIT\cite{DRIT}. 
    
    
\end{itemize}

\begin{figure*}
\begin{center}
\includegraphics[width=1.0 \linewidth]{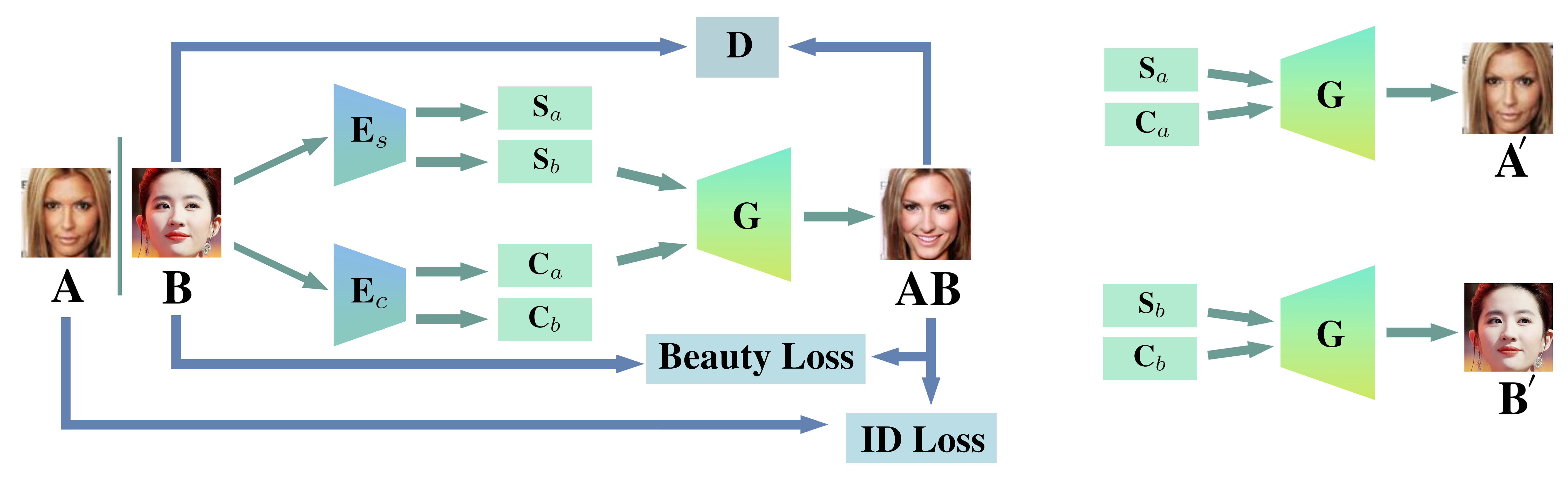}
\end{center}
   \caption{Overview of the proposed network architecture.}
\label{fig:overview}
\end{figure*}

%% file: Section/RelatedWork.tex
\label{sec:related}



\textbf{Makeup and Style Transfer}.
Two recent works on face beauty are BeautyGAN \cite{li2018beautygan} and BeautyGlow \cite{chen2019beautyglow}. In BeautyGlow \cite{chen2019beautyglow}, the makeup features (e.g., eyeshadows and lip
gloss) are first extracted from reference makeup images and then transferred to source non-makeup images. The magnification parameter in
the latent space can be tuned to adjust the extent of the makeup. In BeautyGAN \cite{li2018beautygan}, the issue of extracting/transferring local
and delicate makeup information was addressed by incorporating both global domain-level loss and local instance-level loss in an dual
input/output GAN. 

Face beautification is also related to more general image-to-image translation. Both symmetric (e.g., CycleGAN \cite{zhu2017unpaired}) and asymmetric (e.g., PairedCycleGAN \cite{chang2018pairedcyclegan}) have been studied in the literature; the latter was shown effective for makeup application and removal. Extensions of style transfer into multimodal domain (i.e., one-to-many translations) have been considered in  
MUNIT \cite{huang2018multimodal} and
DRIT \cite{DRIT}.  It is also worth mentioning face image synthesis via StyleGAN \cite{karras2019style} which has demonstrated super-realistic performance.

\textbf{Face Beauty Prediction}. 
The perception of facial appearance or attractiveness is a classical topic in psychology and cognitive sciences \cite{thornhill1999facial,perrett1998effects,perrett1999symmetry}. However, developing 
a computational algorithm that can automatically predict beauty scores from facial images is only a recent endeavor \cite{eisenthal2006facial,gan2014deep}. Thanks to the public release of face beauty database such as SCUT-FBP \cite{xie2015scut}, there has been a growing interest in machine learning based approaches toward face beauty prediction \cite{fan2017label,xu2017facial}.

%% file: Section/ProposedMethod.tex
\begin{figure}[t]
\begin{center}
\includegraphics[width=1.0 \linewidth]{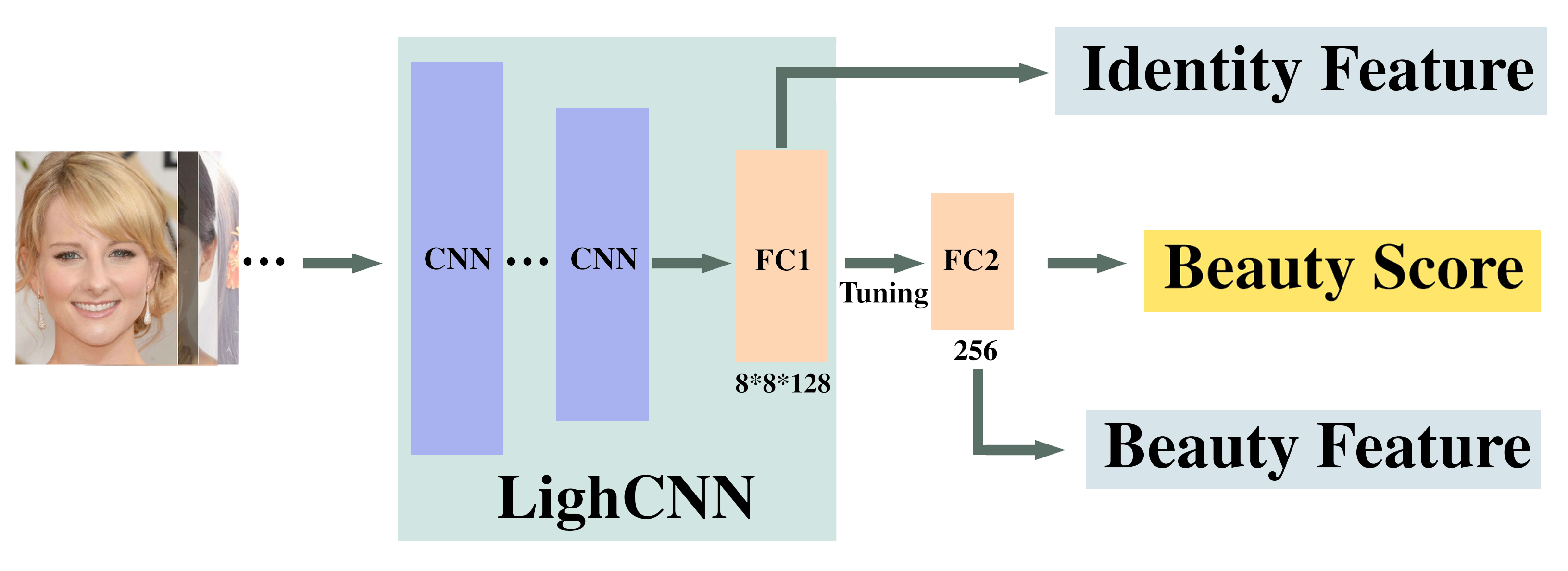}
\end{center}
   \caption{Fine-tuning network for beauty score prediction.}
\label{fig:beauty}
\end{figure}

\subsection{Facial Attractiveness Theory}
\label{sec:discription}

Why facial attractiveness matters? From an
evolutionary perspective, a plausible working hypothesis is that the psychological
mechanisms underlying primates' judgments about attractiveness are consequence of long-period evolution and adaptation. More specifically, facial attractiveness is beneficial to choosing a mate which in turn facilitates the gene propagation \cite{thornhill1999facial}. At the primitive level, facial attractiveness is hypothesized to reflect information about
an individual’s health. Accordingly, conventional wisdom in facial attractiveness research has focused on ad-hoc attributes such as facial symmetry and averageness as potential biomarkers. In the history of modern civilization, the social norm of facial attractiveness has constantly evolved and varies from region to region (e.g., the sharp contrast between eastern and western culture \cite{cunningham1986measuring}).

In particular, facial attractiveness for young females is a stimulating topic as witnessed by the long-lasting popularity of beauty pageants.
In \cite{cunningham1986measuring}, the relation between female facial features and the responses of males was investigated. Based on the male
subjects' attractiveness ratings, two classes of facial features (e.g., large eyes, small nose, and small chin; prominent cheekbones and narrow cheeks) are positively correlated with attractiveness ratings. It is also known from the same study \cite{cunningham1986measuring} that facial features can also predict personality attributions and altruistic inclinations. We opt to focus on face beautification for females only in this work.

\subsection{Problem Formulation and Motivation}
\label{sec:formulation}

Given a target face (an ordinary that is less attractive) and a reference face (usually a celebrity one with a high beauty score), how can we beautify the target face by transferring relevant information from the reference image?
Such problem of face beautification can be formulated as two subproblems: {\em style transfer} and {\em beauty prediction}. Meantime, an important new insight brought into our problem formulation is that the treatment of face beautification as a sequential process where the beauty score of the target face can be gradually improved by a consecutive style transfer steps. As the fine-granularity style transfer proceeds, the beauty score of the beautified target face will monotonically approach that of the reference face. 

The problem of style transfer has been extensively studied in the literature which dated back to content-style separation \cite{tenenbaum2000separating}. The idea of extracting style-based representation (style code) has attracted increasingly more attention in recent years -e.g., \cite{huang2017arbitrary,huang2018multimodal,lee2018diverse,donahue2017semantically,mathieu2016disentangling}. Note that makeup transfer only represents a special case where style is characterized by local features only (e.g., eye-shadow and lip-stick). In this work we conceive a more generalized solution to transfer both global and local style codes from the reference image. The extraction of style codes will be based on the solution to the other problem of beauty prediction. Such sharing of learned features between style transfer and beauty prediction allows us to achieve the fine-granularity control over the process of beautification.


\subsection{Architecture Design}
\label{sec:arch}

As illustrated in Fig. \ref{fig:overview}, we use 
$A$ and $B$ to denote the target face (unattractive) and the reference face (attractive) respectively. The objective of beautification is to translate image $A$ into a new image $AB$ whose beauty score is $Q$-percent close to that of $B$ ($Q$ is an integer between 0 and 100 specifying the granularity of beauty transfer).
Assume both images $A$ and $B$ can be decomposed into a two-part representation consisting of style and content. That is, both images will be encoded by a pair of encoders: content (identity) encoder $E_{c}$ and style (beauty) $E_{s}$ encoder respectively. In order to transfer the beauty style from reference $B$ to target $A$, it is natural to concatenate the content(identity)-based representation $C_a$ with the style(beauty)-based representation $S_b$; and then reconstruct the beautified image $AB$ through a dedicated decoder $G$ defined by 
\begin{equation}
    G(AB) = G[E_{c}(A),E_{s}(A) + E_{s}(B)]. 
\label{eq:1}
\end{equation}
The rest of our architecture in Fig. \ref{fig:overview} mainly includes two components: a GAN-based module ($G$ pairs with $D$) responsible for style transfer and a module of beauty and identity loss responsible for beauty prediction (please refer to Fig. \ref{fig:beauty}). 

Our GAN module consisting of two encoders, one decoder, and one discriminator aims at distilling the beauty/style representation from the reference image and embedding it into the target image for the purpose of beautification. Inspired by recent work \cite{ulyanov2017improved}, we propose to integrate an Instance-Normalization (IN) layer after convolutional layers as part of the encoder for content feature extraction. Meantime, a global average pooling and a fully connected layer follow convolutional layers as part of the encoder for beauty feature extraction. Note that we skip IN in beauty encoder because IN would remove the characteristics of original feature representing critical beauty-related information \cite{huang2017arbitrary} (that's why we keep it within content encoder). To cooperate with beauty encoder and speed up the translation, the decoder is equipped with an Adaptive Instance Normalization (AdaIN) \cite{huang2017arbitrary}. Additionally, we have adopted the popular multi-scale discriminators \cite{Wang_2018_CVPR} with Least-Square GAN (LSGAN) \cite{mao2017least} as the discriminator in our GAN module. 

Our beauty prediction module is based on fine-tuning an existing LightCNN \cite{wu2018lightCNN} as shown in Fig \ref{fig:beauty}. Since it's difficult to train a deep neutral network for beauty prediction from the scratch, we opt to work with LightCNN \cite{wu2018lightCNN} - a pre-trained model for face recognition with millions of face images. Instead, we employ a fine-tuning layer (FC2) to adapt it for beauty score prediction (FC2 plays the role of beauty feature extractor). Meantime, in order to preserve the identity during face beautification, we propose to take the full advantage of our beauty prediction model by piggybacking the identity feature it produced. More specifically, identity feature is generated from the second fully connected layer (FC1) of LightCNN; note that we have only fine-tuned the last fully connected (FC2) for beauty prediction. By using this piggyback trick, we manage to extract both identity and beauty features from one off-shelf model.

\subsection{Fine-granularity Beauty Adjustment}
\label{sec:dyn weight}

As we argued before, beautification should be modeled by a continuous process instead of a discrete domain transfer. In order to achieve the fine-granularity control of the beautification process, we propose to formulate a weighted beautification equation by 
\begin{equation}
    G(AB) = G[E_{c}(A), w_{1}  E_{s}(A) + w_{2}  E_{s}(B)],
\label{eq:2}
\end{equation}
where $w_{1}+w_{2}=1$ and ,$0\leq w_{1},w_{2}\leq 1$. It is easy to observe the two extreme cases: 1) Eq. \eqref{eq:2} degenerates into reconstruction when $w_{1}=1,w_{2}=0$; 2) Eq. \eqref{eq:2} corresponds to the fullest-extent beautification when $w_{1}=0,w_{2}=1$. Such  linear weighting strategy represents a simple solution to adjust the amount of beautification. 

\begin{figure*}
\begin{center}
\includegraphics[width=1.0 \linewidth]{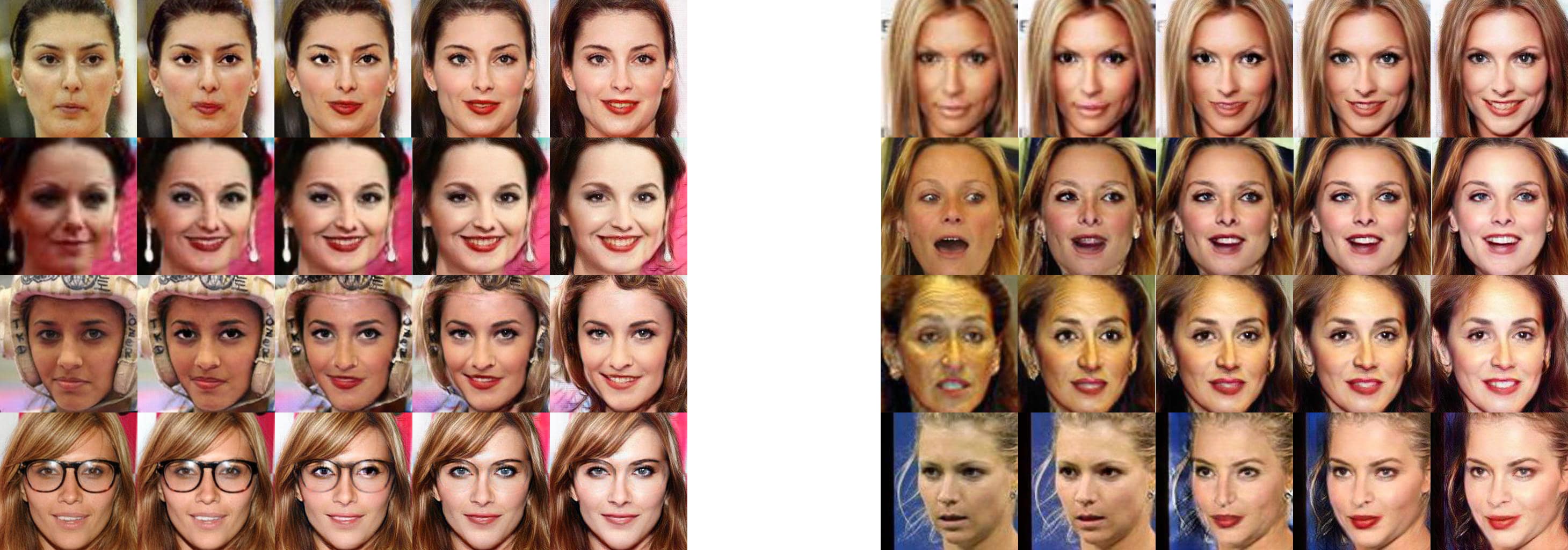}
\end{center}
\caption{Beauty degree adjustment by controlled beauty representation (the leftmost is the original input, from left to right: light to heavy beautification).}
\label{fig:adjustment}
\end{figure*}

To make our model more robust, we have adopted the following training strategy: replacing    $G[E_{c}(A),E_{s}(A)+E_{s}(B)]$ with $G[E_{c}(A), E_{s}(B)]$ in the training stage so that we do not need to train multiple weighted models when weights vary. Instead we apply the weighted beautification equation of Eq. \eqref{eq:2} for testing directly. In other words, we pretend the beauty feature of the target image $A$ is forgotten during the training but partially exploit it during the testing (since it is less relevant than identity feature). In summary, our fine-granularity beauty adjustment strategy heavily counts on the capability of beauty encoder $E_{s}$ for reliably extracting beauty representation.
The effectiveness of the proposed fine-granularity beauty adjustment can be justified by referring to Fig. \ref{fig:adjustment}.

\subsection{Loss Functions}
\label{sec:loss}


\textbf{Image reconstruction}.  Both encoder and decoder need to make sure that target and reference images can be approximately reconstructed from the extracted content/style representation. Here we have adopted $L_1$-norm for reconstruction loss for the reason of robustness.
\begin{align}
	\mathcal{L}^{A}_{\text{REC}} =\mathbb{E}_{a\sim p(a)}[||G[E_s(A),E_c(A)]-A||_{1}] , \nonumber \\
	\mathcal{L}^{B}_{\text{REC}} =\mathbb{E}_{b\sim p(b)}[||G[E_s(B),E_c(B)]-B||_{1}] 
\end{align}
where $||\cdot||_1$ denotes the $L_1$-norm.

\textbf{Adversarial loss}.
We apply adversarial losses \cite{goodfellow2014generative} for matching the distributions of the generated image $AB$ and the target data $B$. In other words, the adversarial loss ensures the beautified face looks as realistic as the reference. 

\begin{align}
\mathcal{L}^{AB}_{\text{GAN}}= \mathbb{E}_{AB}[\log(1-D(G(AB))] 
 + \mathbb{E}_{B}[\log D(B)],
 \label{eq:4}
\end{align}
 where $G(AB)$ is defined by Eq. \eqref{eq:1}.

\textbf{Identity preservation}. To preserve the identity information during the process of beautification, we propose to adopt an identity loss function from the off-shelf face recognition model LightCNN \cite{wu2018lightCNN} trained on millions of faces. Identity features are extracted from the FC1 layer, which is a $2^{13}$-dimensional vector.

\begin{align}
	\mathcal{L}^{A}_{\text{ID}} = ||f_{id}(G[E_{c}(A),E_{s}(A)])-f_{id}(A)||_{1} ,
	\nonumber \\
	\mathcal{L}^{B}_{\text{ID}} = ||f_{id}(G[E_{c}(B),E_{s}(B)])-f_{id}(B)||_{1} ,
	\nonumber \\
	\mathcal{L}^{AB}_{\text{ID}} = ||f_{id}(G[E_{c}(A),E_{s}(B)])-f_{id}(A)||_{1} ,
	\label{eq:5}
\end{align}
where
$\mathcal{L}^{A}_{\text{ID}}$ and $\mathcal{L}^{B}_{\text{ID}}$ are responsible for identity preservation, and $\mathcal{L}^{AB}_{\text{ID}}$ aims at preserving the identity after beautification. Note that our objective is to preserve the identity but improve the beauty in the generated image $AB$ as jointly constrained by Eqs. \eqref{eq:4} and \eqref{eq:5}.

\textbf{Beauty loss}.  In order to leverage the beauty feature from the reference, a beauty prediction model is first used to extract beauty features and then we propose to minimize the $L_1$ distance between the beautified face $AB$ and $B$ as following:

\begin{align}
	\mathcal{L}^{A}_{\text{BT}} = ||f_{bt}(G[E_{c}(A),E_{s}(A)])-f_{bt}(A)||_{1} ,
	\nonumber \\
	\mathcal{L}^{B}_{\text{BT}} = ||f_{bt}(G[E_{c}(B),E_{s}(B)])-f_{bt}(B)||_{1} ,
	\nonumber \\
	\mathcal{L}^{AB}_{\text{BT}} = ||f_{bt}(G[E_{c}(A),E_{s}(B)])-f_{bt}(A)||_{1} ,	
\end{align}
where $f_{bt}$ denotes the operator extracting the 256-dimensional beauty feature (FC2 as shown in Fig. \ref{fig:beauty}).

\textbf{Perceptual loss}. 
Unlike makeup transfer, our face beautification seeks many-to-many mapping in an unsupervised way, which is more challenging especially in view of both inner-domain and cross-domain variations. As mentioned in \cite{ma2018exemplar}, semantic inconsistency is a major issue for such unsupervised many-to-many translation To address this issue, we propose to apply a perceptual loss to minimize the perceptual distance between the beautified face $AB$ and the reference face $B$. This is a modified version from \cite{ma2018exemplar}, where Instance Normalization \cite{ulyanov2017improved}is performed on VGG \cite{simonyan2014very} features before computing the perceptual distance. 

\begin{align}
    \mathcal{L}^{AB}_{\text{P}} = 
    ||f_{vgg}(G(AB))-f_{vgg}(B)||_{2} ,
\end{align}
where $||\cdot||_2$ denotes the $L_2$-norm.




\textbf{Total loss}.
Putting things together, we jointly train the architecture by optimizing the following objective function:

\begin{align}
\underset{E_{c}, E_{s}, G} \min\ \underset{D} \max\ \mathcal{L}(E_{c}, E_{s}, G, D) = \lambda_{1}( \mathcal{L}^{A}_{\text{REC}}  +  \mathcal{L}^{B}_{\text{REC}}) +
	\nonumber \\ 
	\lambda_{2}(\mathcal{L}^{A}_{\text{ID}}+\mathcal{L}^{B}_{\text{ID}}+\mathcal{L}^{AB}_{\text{ID}}) +
	\lambda_{3}(\mathcal{L}^{A}_{\text{BT}}+\mathcal{L}^{B}_{\text{BT}}+\mathcal{L}^{AB}_{\text{BT}}) +
	\nonumber \\
	\lambda_{4}\mathcal{L}^{AB}_{\text{GAN}}+ \lambda_{5}\mathcal{L}^{AB}_{\text{P}} 
	\label{equ:loss}
\end{align}
	where $\lambda_{1}$, $\lambda_{2}$, $\lambda_{3}$, $\lambda_{4}$, $\lambda_{5}$  are regularization parameters.

\begin{figure*}[t]
\begin{center}
\includegraphics[width=0.7 \linewidth]{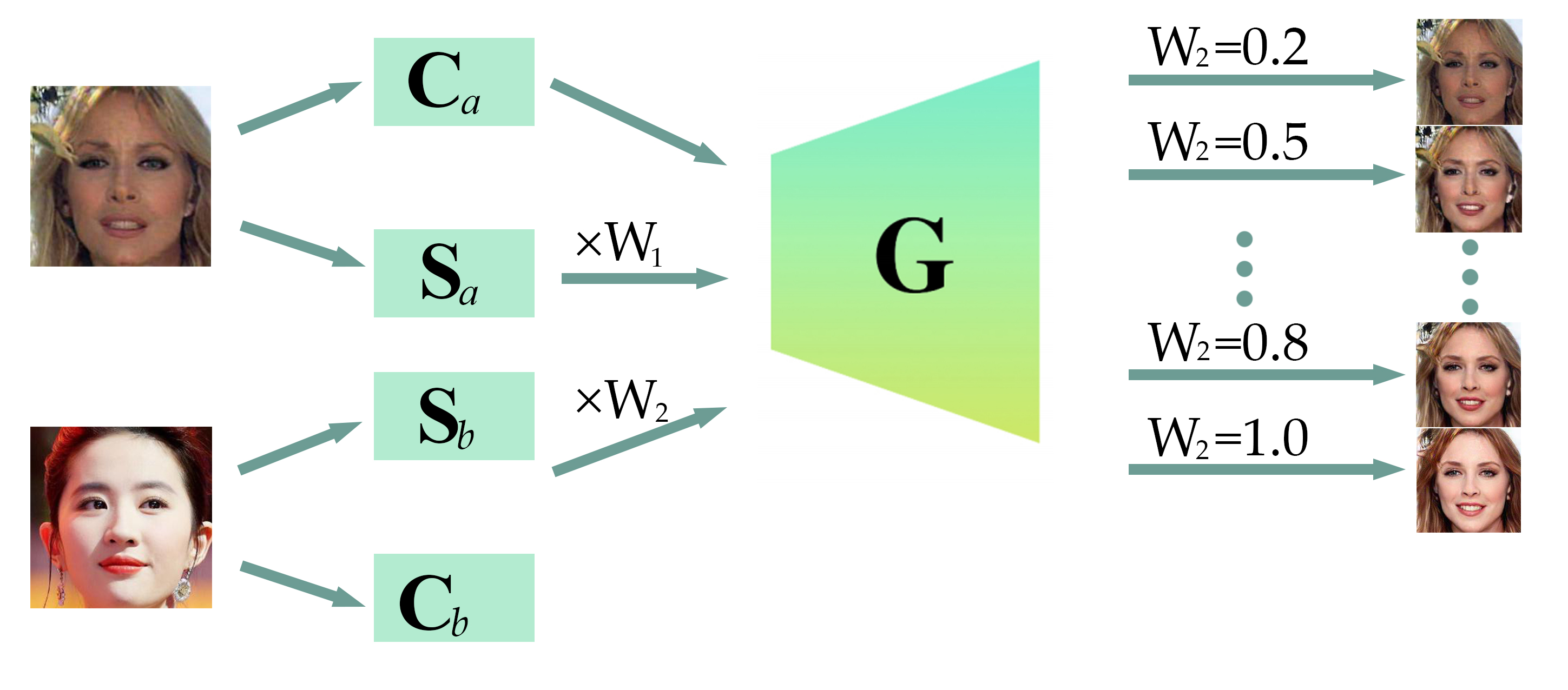}
\end{center}
   \caption{Testing stage for fine-granularity beautification adjustment.}
\label{fig:testing_stage}
\end{figure*}

%% file: Section/Experiments.tex
\begin{figure*}
\begin{center}
\includegraphics[width=1.0 \linewidth]{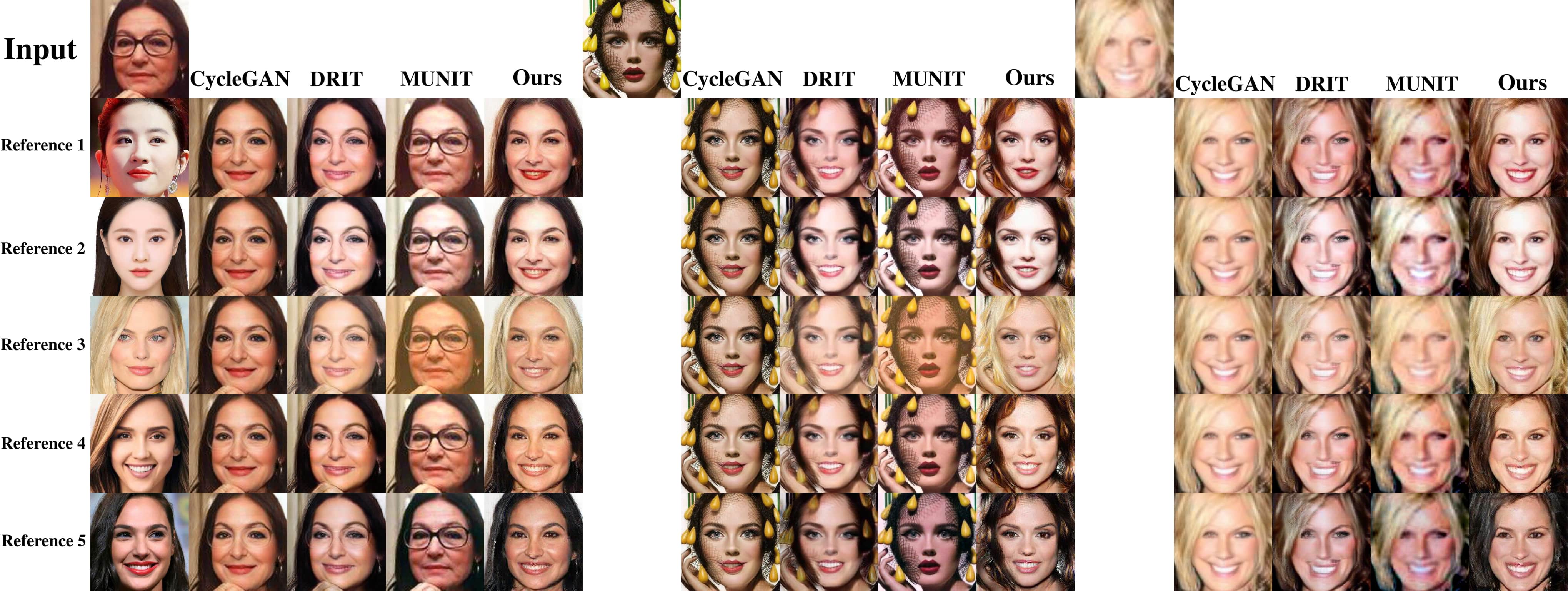}
\end{center}
\caption{Different reference beautification comparison with baseline models. Top images are original input and the left are five references, noted CycleGAN outputs are the same without reference influence.}
\label{fig:compare_ref}
\end{figure*}

\begin{figure*}
\begin{center}
\includegraphics[width=1.0 \linewidth]{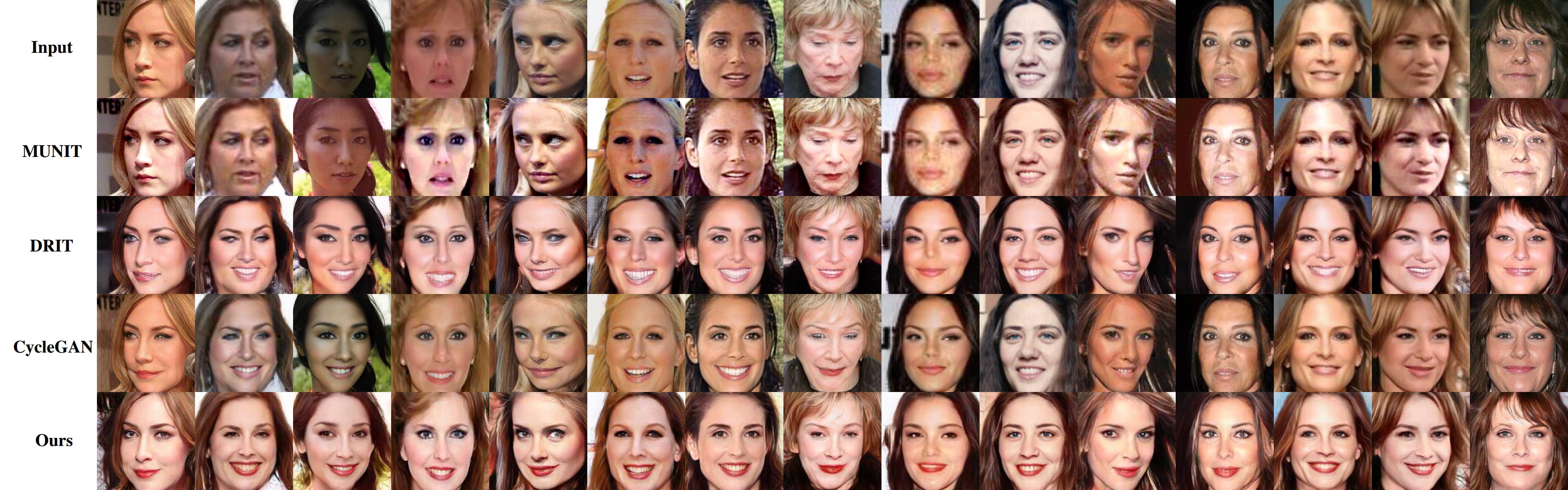}
\end{center}
\caption{Same reference (Reference 1 in Fig \ref{fig:compare_ref}) beautification comparison with baseline models.}
\label{fig:same_ref}
\end{figure*}

\subsection{Training Datasets}
\label{sec:data}

Two datasets are used in our experiments. First, we have used CelebA \cite{liu2015faceattributes} to conduct the beautification experiment (only  female celebrities are considered in this paper).  Authors in \cite{liu2019understanding} have found that some facial attributes have a positive impact on beauty perception. So we have followed their findings to prepare our training datasets - i.e., the images containing those positive attributes (e.g., arched eyebrow, heavy makeup, high cheekbone, wearing lipsticks) as our reference dataset $B$; and images that do not contain those attributes as our target (to be beautified) dataset $A$. 
We have merged the training and validation originate from CelebA as our new training set in order to enlarge the training size, but keep the testing dataset the same as the original protocol \cite{liu2015faceattributes}. Our finalized training set includes 7195 for $A$ and 18273 for $B$,  and testing set has 724 class-$A$ images and 2112 class-$B$ images.  
Another dataset called SCUT-FBP5500 \cite{liang2018scut} is used to train our face beauty prediction network. Following their protocol we have used 60\% samples (3300 images) as training and the rest 40\% (2200) as testing in our experiment.

\subsection{Implementation details}
\label{sec:settings}

\textbf{Generative model}. 
Similar to \cite{huang2018multimodal}, our $E_{c}$ consists of several strided convolutional layers and residual blocks \cite{he2016deep}, all convolutional layers are followed by Instance Normalization (IN) \cite{ulyanov2017improved}. As for $E_{s}$, a global average pooling layer and a fully connected (FC) layer are followed by the strided convolutional layers. IN layer is removed to preserve the beauty features. Inspired by recent GAN works \cite{huang2017arbitrary,dumoulin2016learned,karras2019style} that use affine transformation parameters in normalization layers to better represent style, our decoder $G$ is equipped with the residual blocks as well as Adaptive Instance Normalization (AdaIN). The parameters of AdaIN are dynamically generated by a Multiple Perceptron (MLP) from the beauty codes similar as \cite{huang2018multimodal}, seeing as following:
\begin{equation}
\textrm{AdaIN}(z, \gamma, \beta)= \gamma\left(\frac{z-\mu(z)}{\sigma(z)}\right)+\beta
\end{equation}
where $z$ is the activation of the previous convolutional layer, $\mu$ and $\sigma$ are channel-wise mean and standard deviation, $\gamma$ and $\beta$ are parameters generated by the MLP.

\textbf{Discriminative model}.
We have implemented multi-scale discriminators \cite{wang2018high} to guide generative model to generate both realistic and consistent image in a global view. In addition, LSGAN \cite{mao2017least} is used in our discriminative model to leverage the image quality.  

\textbf{Beauty and identity model}. As shown in Fig. \ref{fig:beauty}, we have used an off-shelf face recognition model-- LightCNN \cite{wu2018lightCNN}, which was trained on millions of faces and achieved state-of-the-art performance in several benchmark studies. In order to extract face beauty feature, we do a fine-tuning based on the pre-trained model from LightCNN,  the last fully connected (FC2) layer is the learnable layer for beauty score prediction and all previous layers are kept fixed during training process. When tested on the popular CUT-FBP5500 dataset \cite{liang2018scut}, our method achieves the MAE of 0.2372 on testing set, which significantly outperforms theirs (0.2518) \cite{liang2018scut} in our experiment.  

In our experimental setting, the off-shelf LightCNN is considered as the identity feature extractor and the fine-tuning beauty prediction model is used as the face beauty extractor. In order to extract both ID and beauty features using one model, we have taken advantage of the beauty prediction model and extract the beauty feature from the last FC layer (FC2 in Fig \ref{fig:beauty}), and the second to last FC layer (FC1 in Fig \ref{fig:beauty}) as the identity feature outputs. When optimization involves two interacting networks, we have found such piggyback idea is more efficient than jointly training both beautification and beauty prediction modules. 




%% file: Section/ResultsEvaluation.tex
\subsection{Baseline Methods}
\label{sec:baseline}

\textbf{CycleGAN} \cite{zhu2017unpaired} A cycle consistency loss was introduced to facilitate the image-to-image translation, which provides a simple but efficient solution to style transfer from unpaired data. 

\textbf{DRIT} \cite{lee2018diverse} An architecture projects images onto two spaces: a domain-invariant content space capturing shared information across domains and a domain-specific attribute space. Similar to CycleGAN, a cross-cycle consistency loss based on disentangled representations is introduced to deal with unpaired data. Unlike CycleGAN, DRIT is capable of generating diverse images on a wide range of tasks.

\textbf{MUNIT} \cite{huang2018multimodal} A framework for multimodal unsupervised image-to-image translation, where images are decomposed into a content code that is domain-invariant and a style code that captures domain-specific properties.  By combining content code with a random style code, MUNIT can also generate diverse outputs from the target domain.

As mentioned in Section \ref{sec:related}, all baseline methods have their weakness when applied to reference-based beautification. CycleGAN cannot take advantage of specific references for translation, the outputs lack diversity once training done. DRIT and MUNIT are capable of many-to-many translation but fail to generate a sequence of correlated images (e.g., faces with increasing beauty scores). By contrast, our model is capable of not only beautifying faces based on a given reference but also controlling the degree of the beautification to fine-granularity, as shown in Fig \ref{fig:adjustment}. 

\subsection{Qualitative and Quantitative Evaluations}
\label{sec:qualitative}

\textbf{User study}. To evaluate the image quality from human's perception, we develop a user study and ask users to vote the most attractive one among ours and the baseline. 100 face images from testing set are submitted to Amazon Mechanical Turk (AMT), and each survey requires 20 users. We collect 2000 data points in total to evaluate human preference. The final results demonstrate the superiority of out model, showing in Table \ref{tab:user_study}. 

\begin{table}[t]
    \centering
    \begin{tabular}{c|c|c}
    \hline \hline
         $\textbf{Model}$ & $\textbf{Count}$ & $\textbf{Percent}$  \\
    \hline
         CycleGAN   &  401 & 20.05 \\
         DRIT       &  282    & 14.1   \\
         MUNIT      &  390    & 19.5   \\
         Ours       &  \textbf{927}    & \textbf{46.35}   \\ 
         \hline
    \end{tabular}
    \caption{User study preference for beautified images.}
    \label{tab:user_study}
\end{table}

\textbf{Beauty Score Improvement}. To further evaluate the effectiveness of the proposed beautification approach, we have fed the beautified images into our face beauty prediction model to output the beauty scores. The beauty prediction model is trained on SCUT-FBP as mentioned before and the scale of beauty score is 5 in that dataset. After calculating and averaging the testing images (724), our model outperforms all other methods and gains a $37.11\%$ increase when compared to average beauty score of the original input as shown in Table \ref{tab:beauty_score}. 

\begin{table}[t]
    \centering
    \begin{tabular}{c|c|c}
    \hline \hline
         $\textbf{Model}$ & $\textbf{Beauty Score}$ & $\textbf{Gain}$  \\
    \hline
     Original   &  0.97    &  -   \\
         CycleGAN   &  1.15 & 18.56\% \\
         DRIT       &  1.25    & 28.87\%   \\
         MUNIT      &  1.01    & 4.12\%   \\
        
         Ours       &  \textbf{1.33}    & \textbf{37.11}\%   \\ \hline
    \end{tabular}
    \caption{Average beauty score after beautification.}
    \label{tab:beauty_score}
\end{table}

\subsection{Discussions and Limitations}

When compared against recently developed makeup transfer such as BeautyGAN \cite{li2018beautygan} and BeautyGlow \cite{chen2019beautyglow}, we note that our approach differs in the following aspect. Similar to BeautyGAN \cite{li2018beautygan}, ours assumes the availability of a reference image; but unlike BeautyGAN \cite{li2018beautygan} focusing on local touchup only, ours is capable of transferring both global and local beauty features from the reference to the target. Similar to BeautyGlow \cite{chen2019beautyglow}, ours can adjust the magnification in the latent space; but unlike BeautyGlow \cite{chen2019beautyglow}, ours can improve the beauty score (rather than only increasing the extent of makeup).

\begin{figure}[t]
\begin{center}
\includegraphics[width=1.0\linewidth]{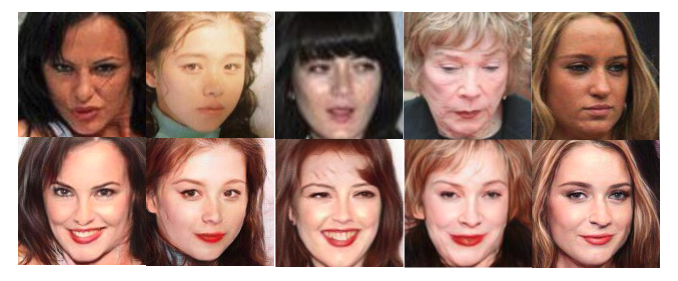}
\end{center}
\caption{Our model is robust to low quality images and small pose variations. }
\label{fig:robust}
\end{figure}

\begin{figure}[t]
\begin{center}
\includegraphics[width=1.0\linewidth]{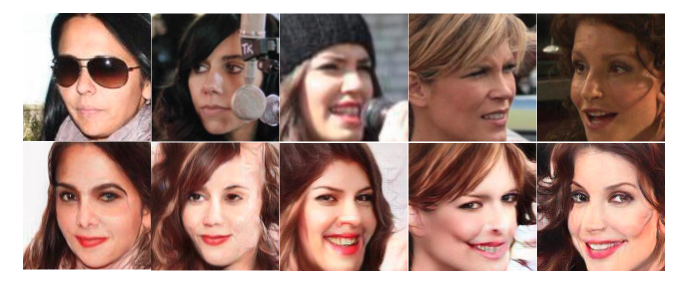}
\end{center}
\caption{Failed case with artifacts: large occlusions and pose variations.}
\label{fig:limit}
\end{figure}

Both user study and beauty score evaluation have demonstrated the superiority of our model. The proposed model is robust to low quality images such as blur and challenging lighting conditions as shown in Fig. \ref{fig:robust}. However, we also notice there are a few typical failed cases in which our model tends to produces noticeable artifacts when the inputs have large occlusions and pose variations (please refer to Fig. \ref{fig:limit}). This is most likely caused by poor alignment - i.e., our references are mostly frontal images; while large occlusion and pose variations lead to misalignment. 

%% file: Section/Conclusion.tex
\begin{figure*}
\begin{center}
\includegraphics[width=1.0 \linewidth]{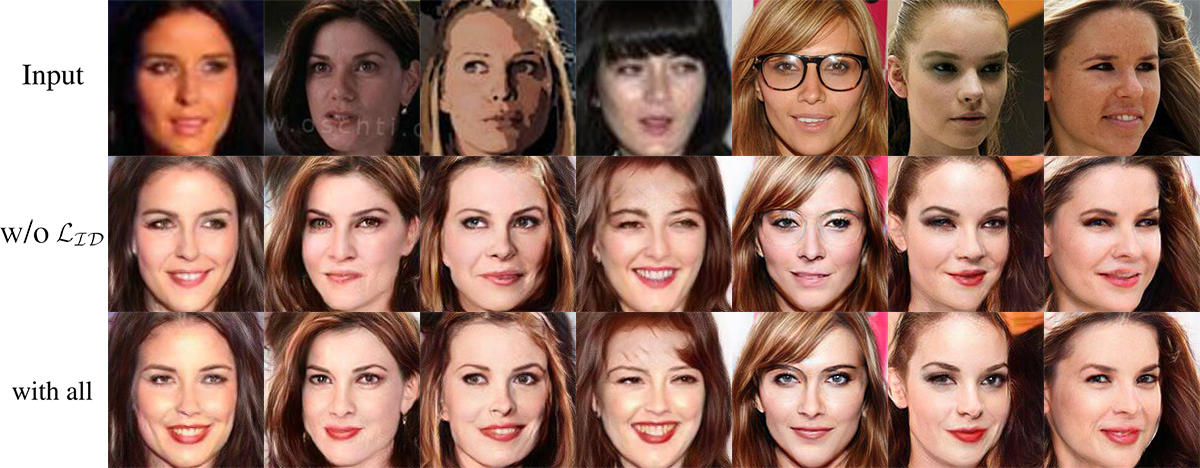}
\end{center}
\caption{Comparisons with and w/o ID Loss $\mathcal{L_{ID}}$ }
\label{fig:ablation_id}
\end{figure*}
\begin{figure*}
\begin{center}
\includegraphics[width=1.0 \linewidth]{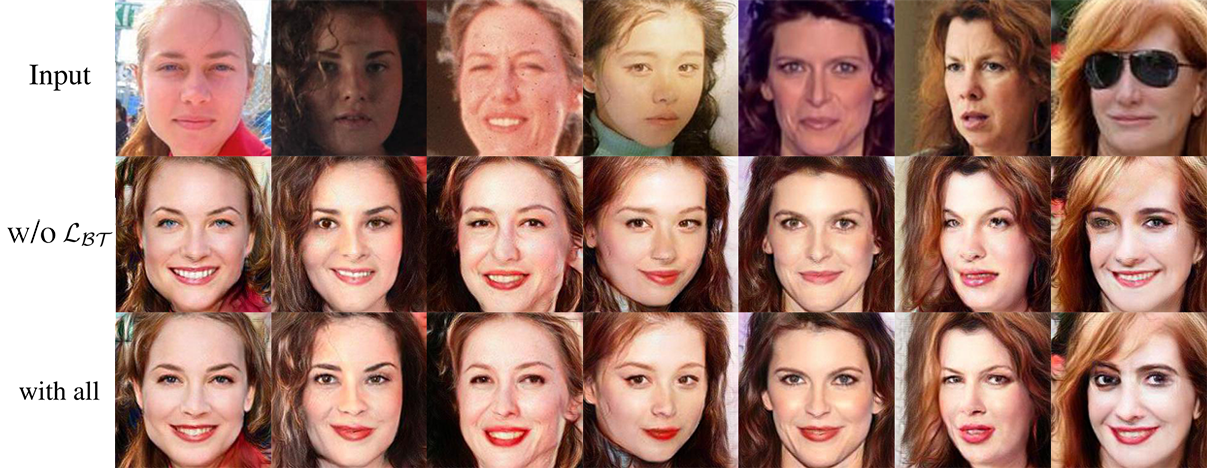}
\end{center}
\caption{Comparisons with and w/o Beauty Loss $\mathcal{L_{BT}}$ }
\label{fig:ablation_beauty}
\end{figure*}
\begin{figure*}
\begin{center}
\includegraphics[width=1.0 \linewidth]{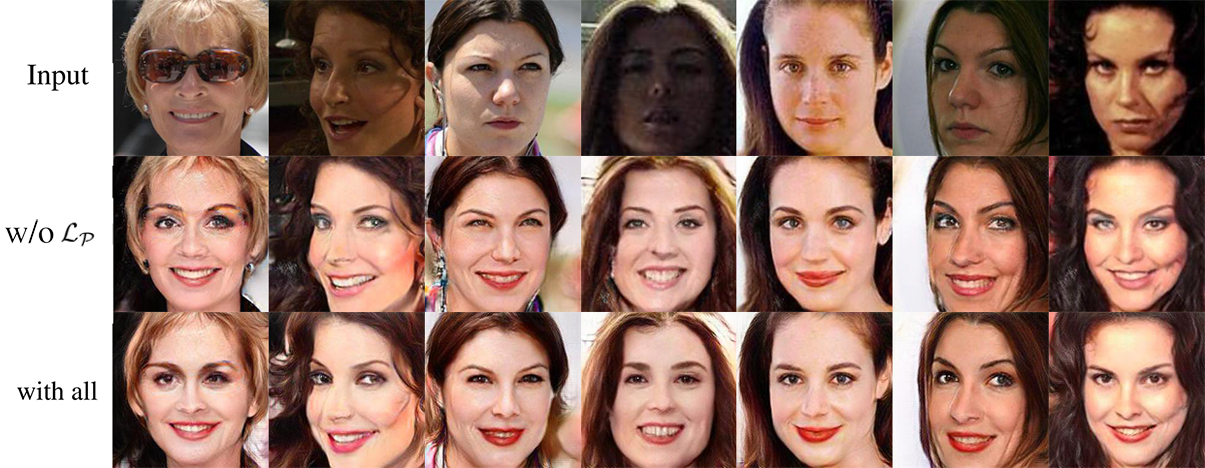}
\end{center}
\caption{Comparisons with and w/o Perceptual Loss $\mathcal{L_{P}}$}
\label{fig:ablation_vgg}
\end{figure*}

In this paper, we have studied the problem of face beautification and presented a novel framework that is more flexible than makeup transfer. Our approach integrates style-based synthesis with beauty score prediction by piggybacking a LightCNN with an GAN-based architecture. Unlike  makeup  transfer,  our  approach targets at many-to-many (instead of one-to-one) translation where  multiple  outputs  can  be  defined  by  either  different references or varying beauty scores. In particular, we have constructed two interacting networks for beautification and beauty prediction. Through a simple weighting strategy, we manage to demonstrate the fine-granularity control of beautification process. Our experimental results have shown the effectiveness of the proposed approach both subjectively and objectively. 

Personalized beautification is expected to attract increasingly more attention in the incoming years. This work we have only focused on the beautification of female Caucasian faces. A similar question can be studied for other populations even though the relationship between gender, race, cultural background and the perception of facial attractiveness has remained under-researched in the literature. How can AI help reshape the practice of personal makeup and plastic surgery is an emerging field for future research.

%% file: Section/Supplemental.tex



\subsection{Ablation Study}

To investigate the importance of each loss, we experiment three variants of our model by removing $\mathcal{L_{ID}}$, $\mathcal{L_{BT}}$ and  $\mathcal{L_{P}}$, one at a time. See Fig~\ref{fig:ablation_id}, \ref{fig:ablation_beauty} and \ref{fig:ablation_vgg} for visual comparisons. These losses compliment each other and work in harmony to reach the optimum beautification effect. This further demonstrates that our loss functions and architecture are well-designed for the facial beautification task.   



\subsection{Network Architectures and Hyperparameters}

\textbf{Generator Architecture.} We adopt our architecture from MUNIT \cite{huang2018multimodal}. Following the convention used in Johnson et al.'s Github repositoty \footnote{\url{https://github.com/jcjohnson/fast-neural-style}}, let $c7s1-k$ denote a $7\times7$ convolutional block with $k$ filters and stride 1. $dk$ denotes a $4\times4$ convolutional block with $k$ filters and stride 2. $rk$ denotes a residual block that contrains two $3\times3$ convolutional blocks. $uk$ denotes a $2\times$ nearest-neighbor upsampling layer followed by a $5\times5$ convolutional block with $k$ filters and stride 1. $gap$ denotes a global average pooling layer and $fc$ denotes a fully connected layer. Instance Normalization (IN) \cite{ulyanov2017improved} is in use to the content (ID) encoder and Adaptive Instance Normalization (AdaIN) \cite{huang2017arbitrary} to style (beauty) encoder. And we use ReLU activations for generator.  The generator architecture is as following:

\begin{itemize}
    \item Content encoder $E_{c}$: $c7s1-64$, $d128$, $d256$, $r256$, $r256$, $r256$
    \item Style encoder $E_{s}$: $c7s1-64$, $d128$, $d256$, $d256$, $d256$, $gap$, $fc$
    \item Decoder $G$: $r256$, $r256$, $r256$, $r256$, $u128$, $u64$, $c7s1-3$
\end{itemize}

\textbf{Discriminator Architecture.} For discriminator, we follow CycleGAN's implementation, and use Leaky ReLU with a slop of 0.2 and multi-scale discriminators with 3 scales. 

\begin{itemize}
    \item Discriminator $D$: $d64$, $d128$, $d256$, $d512$
\end{itemize}

\textbf{Hyperparameters.} The batch size is set as 4 with a single 2080Ti GPU. Our total iteration is 360,000 for a total of around 200 epochs. We use Adam Optimization \cite{kingma2014adam} with $\beta_{1} = 0.5$, $\beta_{2} = 0.999$ and Kaiming initialization \cite{he2015delving}. The learning rate is set as 0.0001 with a 0.5 decay rate in every 100,000 iterations. The style codes from $fc$ has 64 dimension and the loss weights are set as: $\lambda_{1} = 10$, $\lambda_{2} = \lambda_{3} = \lambda_{4} = \lambda_{5} = 1$.